# On the Baldwin Effect under Coevolution

Larry Bull

Computer Science Research Centre

Department of Computer Science & Creative Technologies

University of the West of England, Bristol UK

Larry.Bull@uwe.ac.uk

**Abstract**

The potentially beneficial interaction between learning and evolution – the Baldwin effect – has long been established. This paper considers their interaction within a coevolutionary scenario, ie, where the adaptations of one species typically affects the fitness of others. Using the NKCS model, which allows the systematic exploration of the effects of fitness landscape size, ruggedness, and degree of coupling, it is shown how the amount of learning and the relative rate of evolution can alter behaviour.

**Introduction**

Coevolution refers to the effects upon the evolutionary behaviour of one species by the one or more other species with which it interacts. At an abstract level coevolution can be considered as the coupling together of the fitness landscapes of the interacting species. Hence the adaptive moves made one species in its fitness landscape causes deformations in the fitness landscapes of its coupled partners. In this paper Kauffman and Johnsen's [1992] NKCS model, which allows for the systematic alteration of various aspects of a coevolving environment, particularly the degree of landscape ruggedness and connectedness, is used to explore the Baldwin effect [Baldwin, 1896][Lloyd-Morgan, 1896][Osborn, 1896] within a coevolutionary scenario. The Baldwin effect is here defined as the existence of phenotypic plasticity that enables an organism to exhibit a significantly different (better) fitness than its genome directly represents. Over time, as evolution is guided towards such regions under selection, higher fitness alleles/genomes which rely less upon the phenotypic plasticity can be discovered and become assimilated into the population (see [Sznajder et al., 2012] for an overview).

Hinton and Nowlan [1987] were the first to investigate the Baldwin effect, showing that enabling genetically specified neural networks to alter inter-neuron connections randomly during their lifetime meant the evolutionary system was able to find an isolated optimum, something the system without learning struggled to achieve. That is, the ability to learn "smoothed" the fitness landscape into a unimodal hill/peak. They also found that over time more and more correct connections became genetically specified and hence less and less random learning was necessary; the evolutionary process was guided toward the optimum by the learning process. Belew (eg, [1989]) added the Baldwin effect via backpropogation to his work on the evolution of neural networks for various classes of problem, finding that the search process was greatly improved. Stork and Keesing [1991] then showed how both the frequency with which learning was applied and the number of connection weight adjustment iterations used on each learning cycle impacted upon the benefit gained. Their finding was generalized in [Bull, 1999] where it was shown how the most beneficial frequency and amount of learning varies with the ruggedness of the underlying fitness landscape. Whilst learning and evolution have long been used together within coevolutionary scenarios (after [Ackley & Littman, 1992]), no previous systematic exploration is known. Results here show that behavior is often different both from the standard Baldwin effect and coevolution without learning.

**The NKCS Model of Coevolution**

Kauffman and Levin [1987] introduced the NK model to allow the systematic study of various aspects of fitness landscapes (see [Kauffman, 1993] for an overview). In the standard model, the features of the fitness landscapes are specified by two parameters: $N$, the length of the genome; and $K$, the number of genes that has an effect on the fitness contribution of each (binary) gene. Thus increasing $K$ with respect to $N$ increases the epistatic linkage, increasing the ruggedness of the fitness landscape. The increase in epistasis increases the number of optima, increases the steepness of their sides, and decreases their correlation. The model assumes all intragenome interactions are so complex that it is only appropriate to assign random values to their effects on fitness. Therefore for each of the possible $K$ interactions a table of $2^{(K+1)}$ fitnesses is created for each gene with all entries in the range 0.0 to 1.0, such that there is one fitness for each combination of traits. The fitness contribution of each gene is found from its table. These fitnesses are then summed and normalized by $N$ to give the selective fitness of the total genome.

Kauffman and Johnsen [1992] subsequently introduced the abstract NKCS model to enable the study of various aspects of *co*evolution. Each gene is said to also depend upon $C$ randomly chosen traits in each of the other $S$ species with which it interacts. The adaptive moves by one species may deform the fitness landscape(s) of its partner(s). Altering $C$, with respect to $N$, changes how dramatically adaptive moves by each species deform the landscape(s) of its partner(s). Again, for each of the possible $K+(SxC)$ interactions, a table of $2^{(K+(SxC)+1)}$ fitnesses is created for each gene, with all entries in the range 0.0 to 1.0, such that there is one fitness for each combination of traits. Such tables are created for each species (Figure 1).

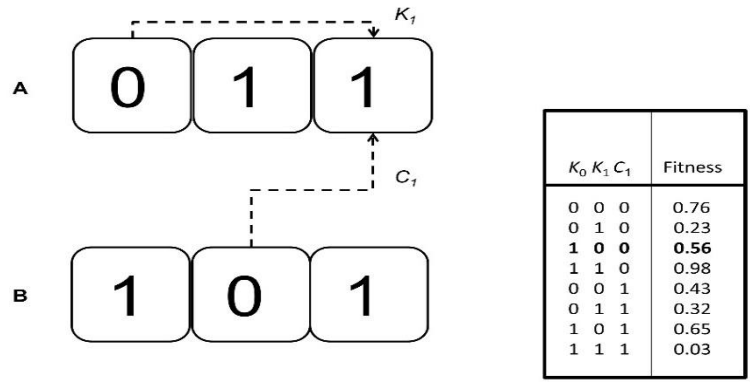

Figure 1: The NKCS model: Each gene is connected to *K* randomly chosen local genes and to *C* randomly chosen genes in each of the *S* other species. A random fitness is assigned to each possible set of combinations of genes. These are normalised by *N* to give the fitness of the genome. Connections and table shown for one gene in one species for clarity. Here *N*=3, *K*=1, *C*=1, and *S*=1.

Kauffman and Johnsen used a mutation-based hill-climbing algorithm, where the single point in the fitness space is said to represent a converged species, to examine the properties and evolutionary dynamics of the NKCS model. That is, each population is of size one and species evolve in turn by making a random change to one randomly chosen gene per generation. The "population" is said to move to the genetic configuration of the mutated individual if its fitness is greater than the fitness of the current individual; the rate of supply of mutants is seen as slow compared to the actions of selection. Ties are broken at random. All results reported in this paper are the average of 10 runs (random start points) on each of 10 NKCS functions, ie, 100 runs, for 20,000 generations. Here 0≤*K*≤10, 0<*C*≤5, for *N*=20 and *N*=100, with *S*=1, ie, two species/populations.

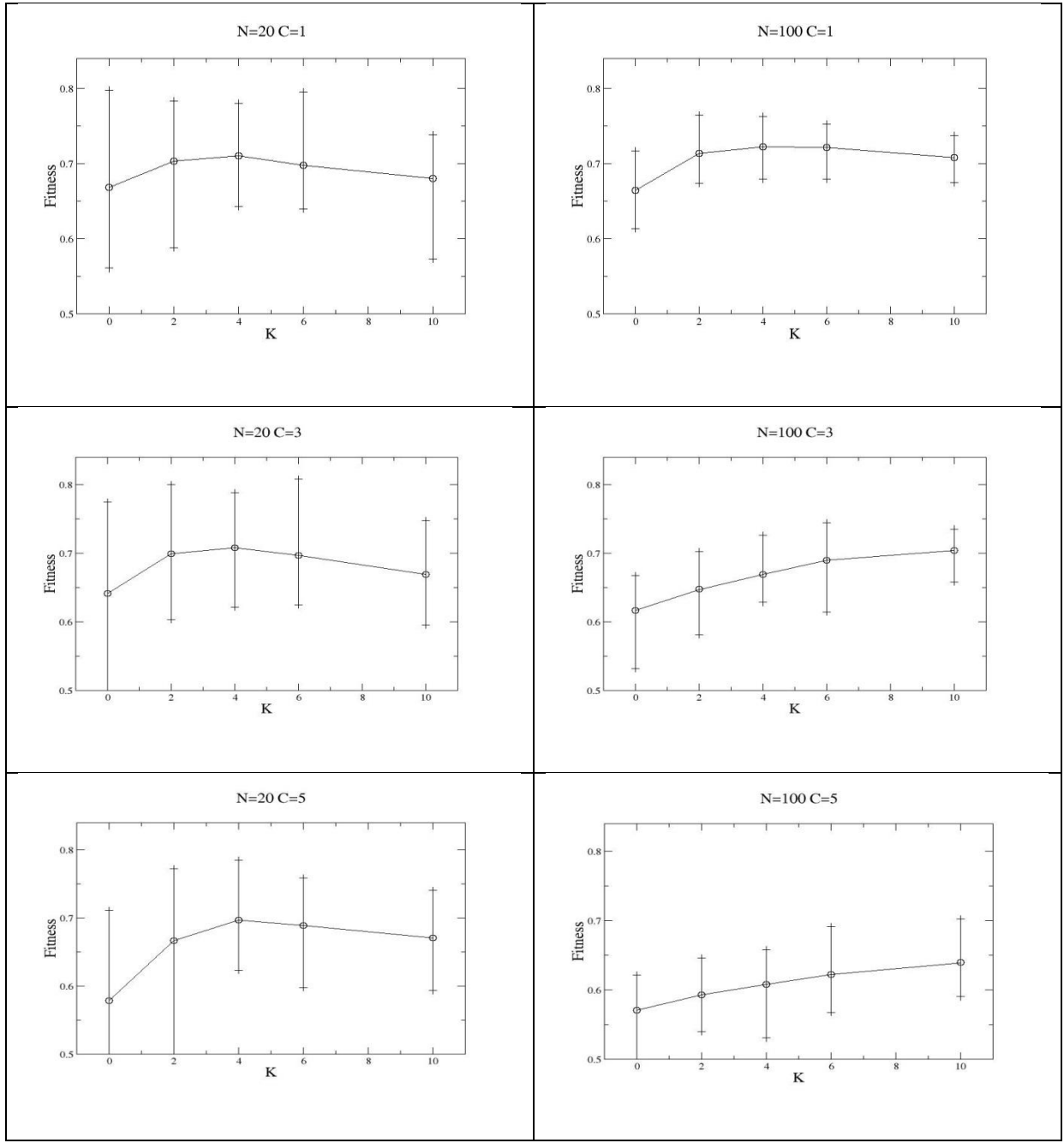

Figure 2: Showing the fitness reached after 20,000 generations on landscapes of varying ruggedness (*K*), coupling (*C*), and length (*N*). Error bars show min and max values.

Figure 2 shows example results for one of two coevolving species where the parameters of each are the same and hence behaviour is symmetrical. When *C*=1, Figure 2 shows examples of the general properties of adaptation on such fitness landscapes mentioned above in the NK model (where *C*=0): at low levels of *K* (0<*K*<8), the landscape buckles up and becomes more rugged, with an increasing number of peaks at higher fitness levels, regardless of *N*. Thereafter the increasing complexity of constraints between genes means the height of peaks typically found begin to fall as *K* increases relative to *N*: for large *N*, the central limit theorem suggests reachable optima will have a mean fitness of 0.5 as $K \rightarrow N$. Figure 2 further shows how increasing the degree of connectedness (*C*) between the two populations causes fitness levels to fall significantly (T-test, $p<0.05$) when $C \geq K$ for $N=20$. That is, as $K \rightarrow N$ a high number of peaks of similar height typically exist in each of the fitness landscapes and so the effects of switching between them under the influence of *C* is reduced since each landscape is very similar. Note this change in behaviour around $C=K$ was suggested as significant in [Kauffman, 1993], where $N=24$ was used throughout. However, Figure 2 also shows how with $N=100$ fitness *always* falls significantly with increasing *C* (T-test, $p<0.05$), regardless of *K*.

**Learning in the NKCS Model**

Following [Bull, 1999], a very simple (random) learning process to enable phenotypic plasticity can be added to evolution by allowing a new individual to make a further *L* (unique) mutations after the first. If the averaged fitness of this "learned" configuration and that of the first mutant is greater than that of the original, the species is said to move to the *first* mutant configuration but assigned the *averaged fitness* of the two configurations.

Figure 3 shows the performance of the Baldwin effect across a range of *K* and *L* combinations. As can be seen for $N=20$, learning of any amount proves beneficial when $K>C$ (T-test, $p<0.05$), with no significant difference in varying *L* (T-test, $p \geq 0.05$). Conversely, when $N=100$, learning is typically either of no benefit or detrimental, particularly for $L>1$. It can be noted that the aforementioned study in the NK model found higher levels of learning were more beneficial with increasing *K* [Bull, 1999]. Hence the underlying dynamics of coevolutionary systems are clearly different, even with $C=1$, where only $L=1$ is beneficial and only for $0<K<10$ (T-test, $p<0.05$). Following [Bull, 1999], varying the frequency with which learning occurs has also been explored with $L=1$, eg, using learning on every other generation or one in four. No significant difference in behaviour to Figure 3 is seen (not shown).

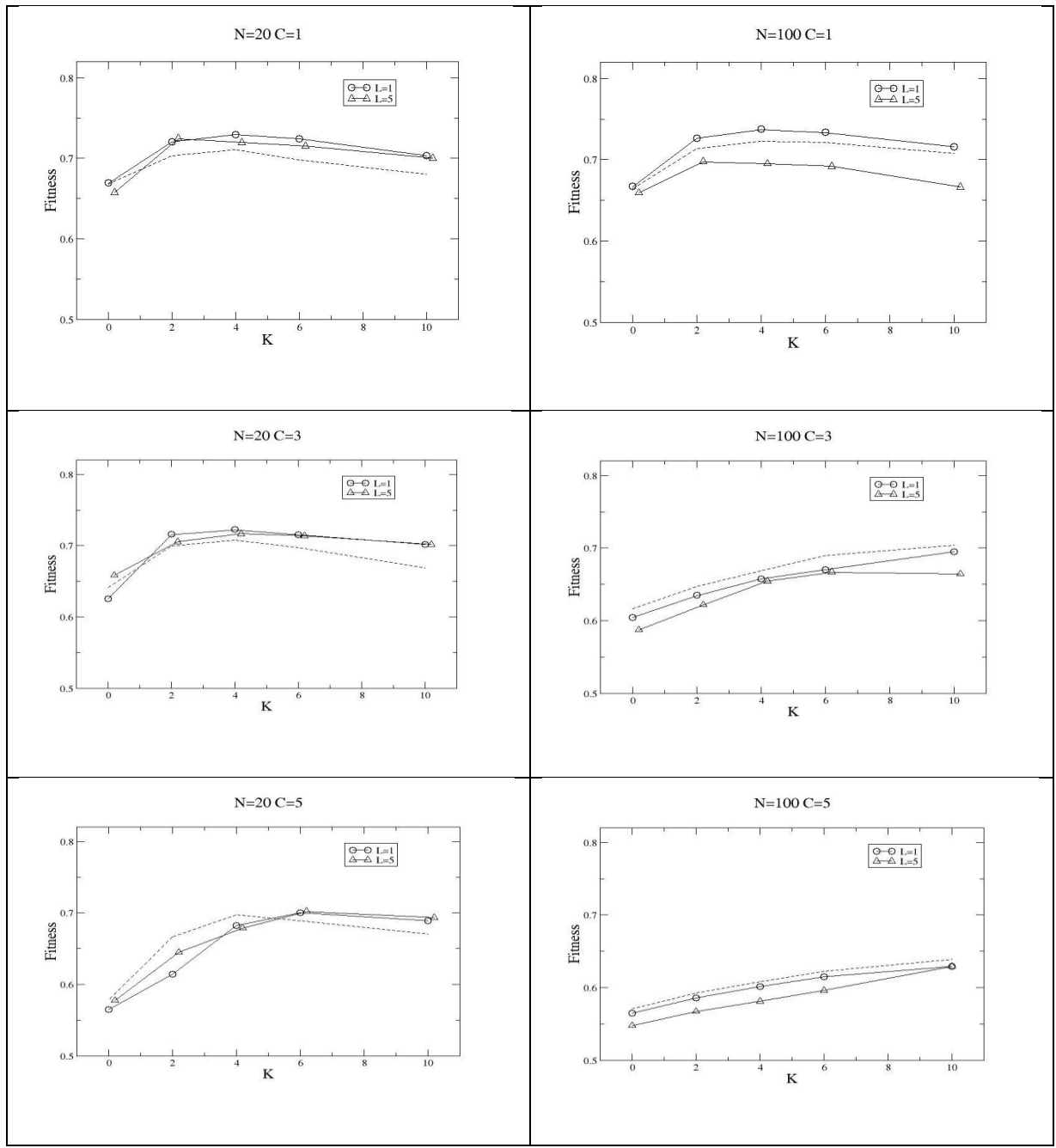

Figure 3: Showing the fitness reached after 20,000 generations for a species using simple learning of varying amounts (*L*) whilst coevolving with a species not using learning, on landscapes of varying ruggedness (*K*), coupling (*C*), and length (*N*). Dotted line shows *L*=0 case (Figure 2). Error bars are not shown for clarity

When both species are using the simple learning mechanism the findings are the same as above for *N*=20, regardless of *L* (T-test, *p*≥0.05). When *N*=100, behaviour also remains the same for *L*=1 but fitness is significantly improved when *L*=5 compared to the above (T-test, *p*<0.05)(not shown).

## Reproduction Rates in the NKCS Model

As in [Kauffman & Johnsen, 1992], the above models have assumed all species coevolve at the same rate; each species coevolves in turn. Following [Bull, et al. 2000], a new parameter $R$ can be added to the model to represent the relative rate at which one species evolves - by undertaking $R$ rounds of mutation and selection to one round in the other(s). Figure 4 shows how with $N$=100, generally, increasing $R$ increases the effects of $C$ for a given $K$ when $C$>1 with -20≤$R$≤20. In contrast, no notable effect is seen by increasing $R$ for any $K$ and $C$ values tried when $N$=20 (not shown). It can be noted that $N$=64 (only) was used in [Bull et al., 2000].

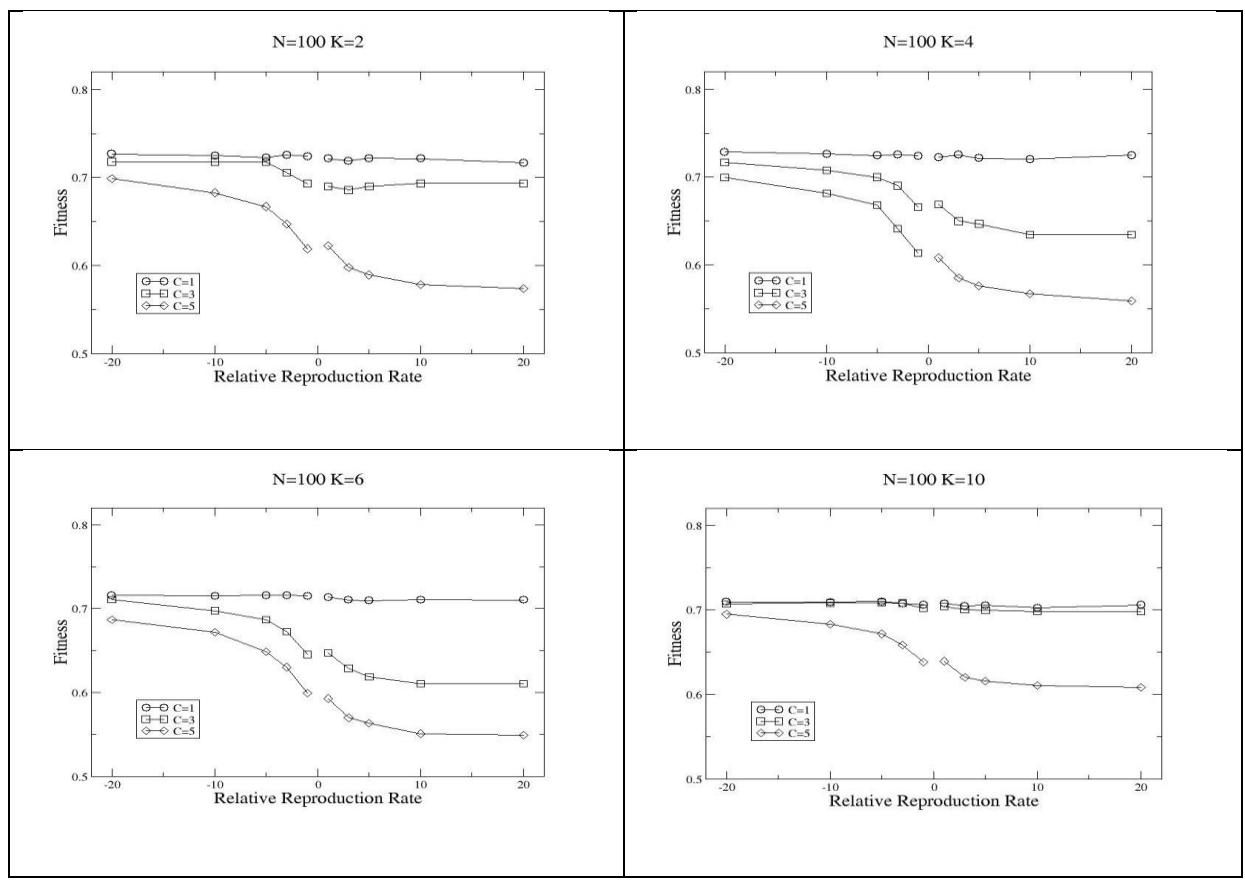

Figure 4: Showing examples of the fitness reached after 20,000 generations for species coevolving with a partner reproducing at a different relative rate $R$. Error bars are not shown for clarity.

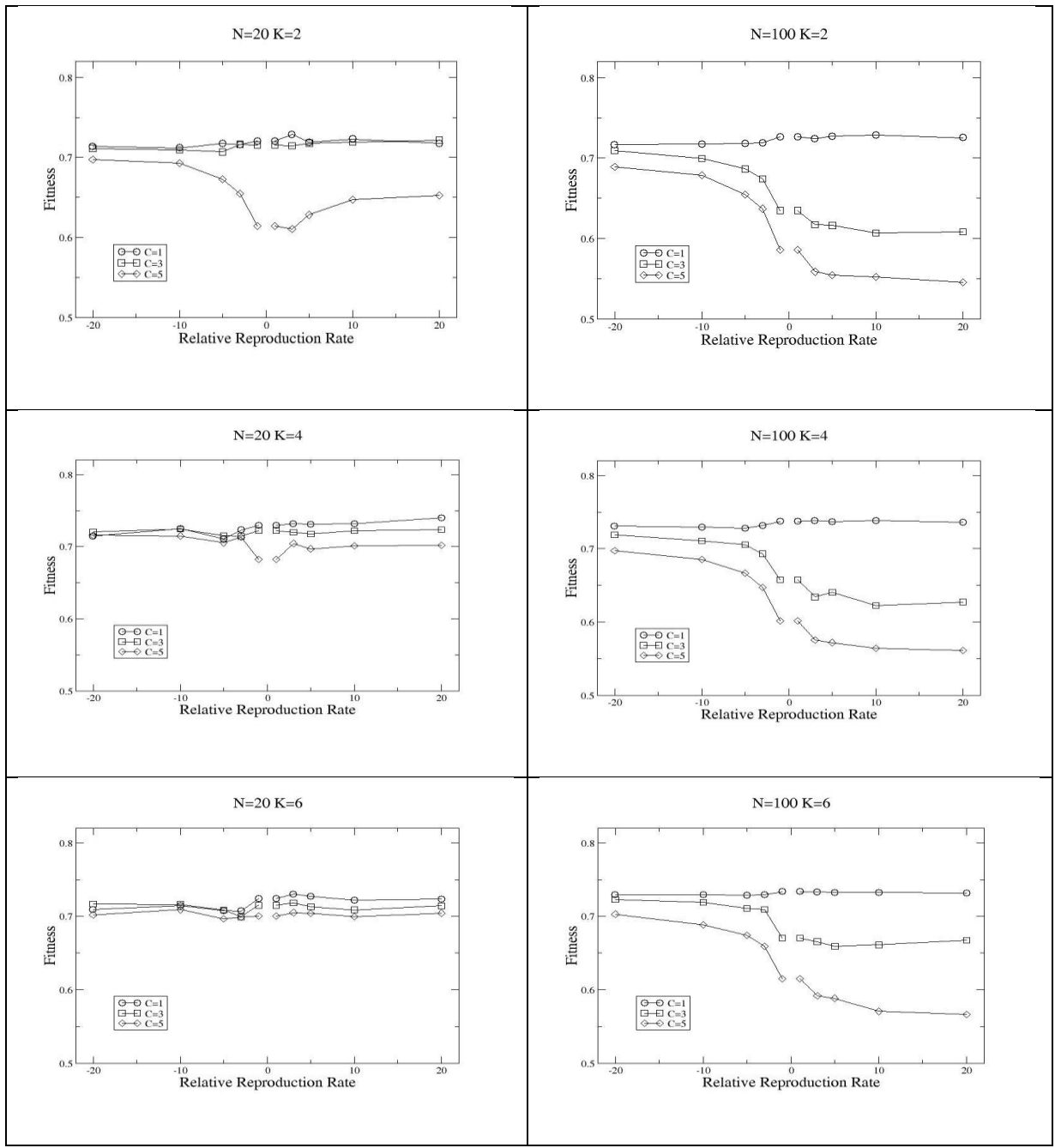

Figure 5: Showing examples of the fitness reached after 20,000 generations for a species using simple learning (*L*=1) coevolving with a partner (without learning) reproducing at a different relative rate *R*.

Figure 5 shows examples of the behaviour of adding learning within such coevolutionary environments, with *L*=1. Typically, there is no significant change in general behaviour. However, with *N*=20 and *C*=5, fitness is seen to *increase* with increasing *R* for *R*>1 when 0<*K*<5 (T-test, *p*<0.05), in contrast to its remaining unchanged without learning. As shown above, learning is beneficial when

$C>K$ with $N=20$ and $R=1$. The improvement in fitness seen here with increasing $R$ is typically not sufficient to improve over the $L=0$ case, except when $K=4$ (Figure 6, left).

Again, varying the amount of learning has also been explored. Results suggest no significant change in fitness is typically seen between $L=1$ and $L=5$ when $N=20$, as $R$ increases (not shown). This is also the case for $N=100$ except for when $C=5$ and $K>6$ where fitness increases with increasing $L$ (T-test, $p<0.05$). This increase in fitness is typically sufficient to negate the effects of increasing $R$, eg, as shown in Figure 6 (right).

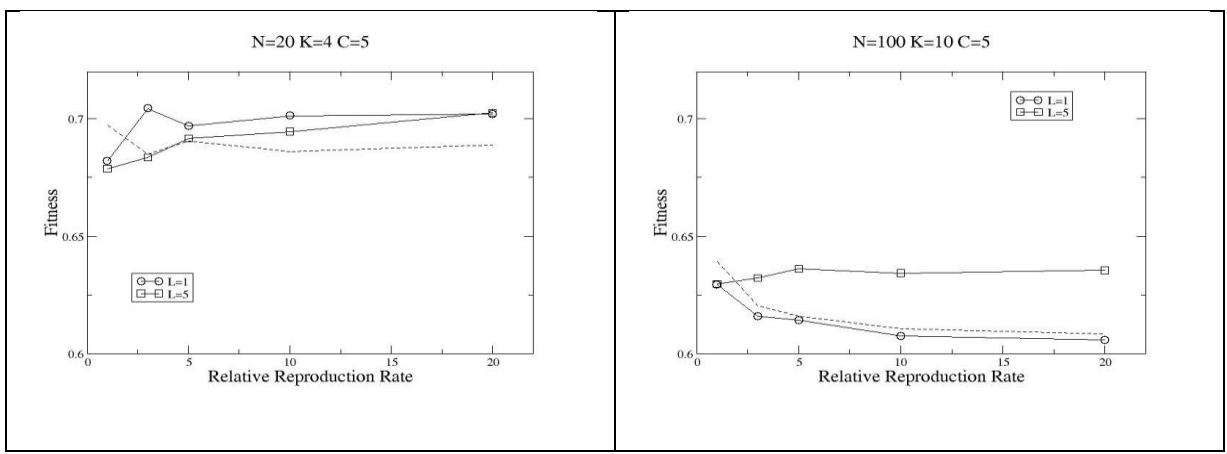

Figure 6: Showing examples of the fitness reached after 20,000 generations for a species using simple learning coevolving with a partner (without learning) reproducing at a different relative rate $R$, for varying amounts of learning ($L$). The dotted line shows the equivalent case without learning.

Having both species use learning has also been explored here. Behaviour is typically the same as above as $R$ increases, with $L=1$ for both $N=20$ and $N=100$ (not shown). Although, fitness drops when $C>K$ with $N=20$ – which results in the loss of the cases of improvement noted above. The same is true for increasing $L$ (not shown).

## Conclusion

Using the NKCS model, this paper has explored the utility of the Baldwin effect in a coevolutionary context. It has been shown how the size, ruggedness, and degree of coupling of the fitness landscapes affects performance. Following [Bull, 1999], this has been shown to be the case when the amount of learning is varied, and, following [Bull et al., 2000], when the species coevolve at different rates.